\documentclass[]{article}
\usepackage{authblk}
\usepackage{listings}
\usepackage{amsmath}
\usepackage{amssymb}
\usepackage{graphicx}
\usepackage{xcolor}
\usepackage{hyperref}
\usepackage{booktabs}
\usepackage{subcaption}
\usepackage{tikz}
\usepackage{float}

\newcommand{\funding}[1]{
	\vspace{6pt}\noindent{\fontsize{9}{9}\selectfont\textbf{Funding:} {#1}\par}}

\definecolor{SourceColor}{RGB}{85,168,104}
\definecolor{TargetColor}{RGB}{221,132,82}
\definecolor{AbsorbingAreaColor}{RGB}{196,78,82}
\definecolor{ObstacleColor}{RGB}{179,179,179}
\definecolor{StairColor}{RGB}{129,114,178}
\definecolor{MeasurementAreaColor}{RGB}{255,0,0}
\definecolor{AgentColor}{RGB}{0,0,255}

\newcommand{\AgentRadius}{0.195000}

\newcommand{\MeasurementAreaOpacity}{0.549020}

\title{Can we learn where people come from? \newline Retracing of origins in merging situations}
\author{Marion G\"odel\footnote{Corresponding author: marion.goedel@hm.edu}, Luca Spataro, Gerta K\"oster}
\affil{Department of Computer Science and Mathematics, \newline Munich University of Applied Sciences \newline Department of Informatics, Technical University of Munich}

\begin{document}

\maketitle

\begin{abstract}
	One crucial information for a pedestrian crowd simulation is the number of agents moving from an origin to a certain target. While this setup has a large impact on the simulation, it is in most setups challenging to find the number of agents that should be spawned at a source in the simulation. Often, number are chosen based on surveys and experience of modelers and event organizers. These approaches are important and useful but reach their limits when we want to perform real-time predictions. In this case, a static information about the inflow is not sufficient. Instead, we need a dynamic information that can be retrieved each time the prediction is started. Nowadays,  sensor data such as video footage or GPS tracks of a crowd are often available. If we can estimate the number of pedestrians who stem from a certain origin from this sensor data, we can dynamically initialize the simulation. 
	In this study, we use density heatmaps that can be derived from sensor data as input for a random forest regressor to predict the origin distributions. We study three different datasets: A simulated dataset, experimental data, and a hybrid approach with both experimental and simulated data. In the hybrid setup, the model is trained with simulated data and then tested on experimental data. The results demonstrate that the random forest model is able to predict the origin distribution based on a single density heatmap for all three configurations. This is especially promising for applying the approach on real data since there is often only a limited amount of data available.
\end{abstract}
\newpage

\section{Introduction}
One crucial initial condition for a pedestrian crowd simulation is the number of agents moving from an origin to a certain target. This setting has a high impact on the simulation result. 
At the same time, it is in most setups challenging to find the number of agents that should be generated at a source in the simulation. Often, number are chosen based on surveys and experience of modelers and event organizers. 
These approaches are important and useful but reach their limits when we want to perform real-time predictions. In this case, a static information about the inflow is not sufficient. Instead, we need an updated information each time the prediction is calculated. 
We want to use available sensor data of the pedestrians currently present in an observation area to estimate the number of pedestrians who stem from a certain origin. This can serve as an approximation of the number of pedestrians entering the scenario at this origin for the next prediction.
As sensors often video cameras are present on site of urban events. Outside of controlled experiments, it is highly challenging to estimate the trajectories of single pedestrians from the video footage. Nevertheless, there are several approaches to estimate the current density from video footage. Consequently, we use density heatmaps as input for this study.
	
\subsection{Research question}
Can we use machine learning methods to estimate the distributions of pedestrians on the origins based on a single density heatmap? 
	
\subsection{Background}
In previous work, we were able to demonstrate that it is possible to prediction direction distributions at crossing based on simulated data with an accuracy  about $80$\% as a proof-of-concept \cite{goedel-2018}. 
 Now, we want to test the methodology on data of pedestrians as e.g., surveillance data, experimental data, or similar. We need trajectories to create the heatmaps and for the response, the actual distributions of the origins. 
 Unfortunately, there are only a few public data sets of pedestrian trajectories available. In particular, to our knowledge there is no public dataset of a crossing scenario with unidirectional flow available. 
 There are, on the other hand, several datasets on merging behavior of pedestrians. Therefore, we decided to test our routines with simulated and experimental data on a merging scenario. In this setup, the distribution on the destination is trivial since everyone heads to the same target. Nevertheless, it is interesting to study if we can find out from which origins pedestrians come. 
 We hope that the problems are similar enough that if the retracing of the pedestrians to the origin distributions is successful, the prediction of the target distributions is also successful on real data. 
 In both cases, the available input data are density heatmaps and we are trying to infer the percentage of pedestrians coming from or heading to a known list of origins or destinations, respectively. 

\section{Methods and materials}
We choose the T-Junction experiments from \cite{boltes-2011,zhang-2011} as scenario. The trajectories as well as video footage of the experiment are available online \footnote{\url{https://doi.org/10.34735/ped.2009.7}}.  

\subsection{Simulation setup}
 All simulations are carried out with the Optimal Steps Model \cite{seitz-2012, sivers-2015} of the open-source simulation framework Vadere \cite{kleinmeier-2019}. We performed a manual calibration of the parameters to the experimental data. In particular, the navigation field was adapted to the data. It is calculated with obstacles method  with factor \texttt{obstacle\-Density\-Weight = 0.3}. That means that the obstacles are regarded in particular way in the calculation of the navigation field. As a result, agents keep more distance to the walls. For the comparison of simulated and experimental data, we used the measurement areas defined in 
 \cite{boltes-2011}.

\begin{figure}[H]
	\resizebox{\textwidth}{!}{%
	\input{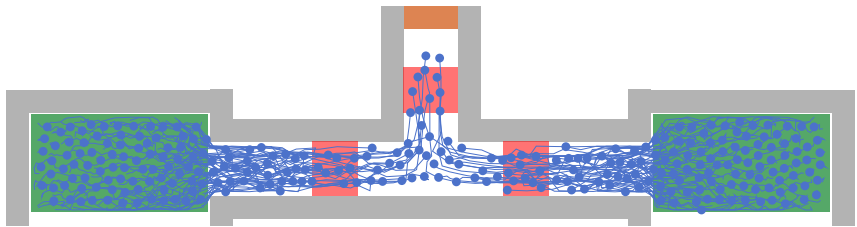}
	}
	\caption{Snapshot of the simulation with Vadere. Agents are moving from two origins (green) to one destination (orange) through a T-junction setup. Measurement areas are shown in red.}
	\label{fig:simulation}
\end{figure}

\subsubsection{Calibration of the simulation: navigation field }
The Voronoi diagrams are calculated with the package \texttt{spatial} of the Python library \texttt{scipy} \cite{virtanen-2020}. 
	
\begin{figure}[H]
		\begin{subfigure}{0.49\linewidth}
			\centering
			\includegraphics[width=\textwidth]{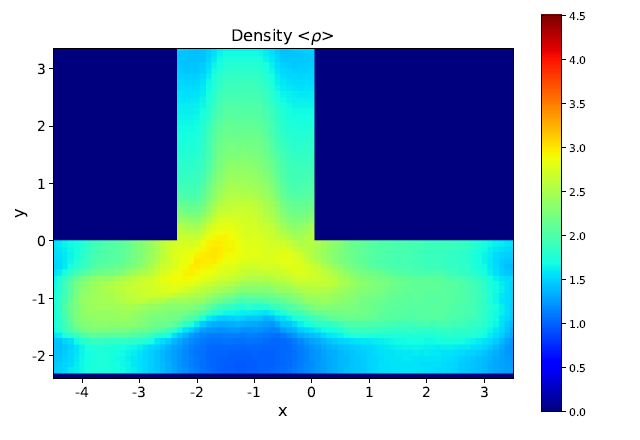}
			\caption{Experimental data set.}
		\end{subfigure} \newline
		\begin{subfigure}{0.49\linewidth}
			\centering
			\includegraphics[width=\textwidth]{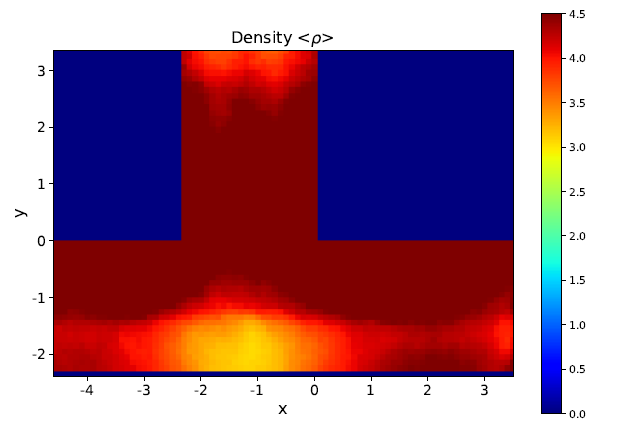}
			\caption{Simulated data set: static floor field. \newline}
		\end{subfigure}
		\begin{subfigure}{0.49\linewidth}
			\centering
			\includegraphics[width=\textwidth]{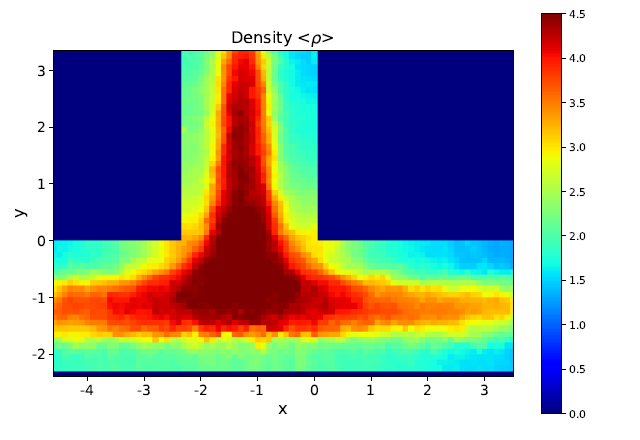}
			\caption{Simulated data set: dynamic floor field.}
		\end{subfigure}
		\begin{subfigure}{0.49\linewidth}
			\centering
			\includegraphics[width=\textwidth]{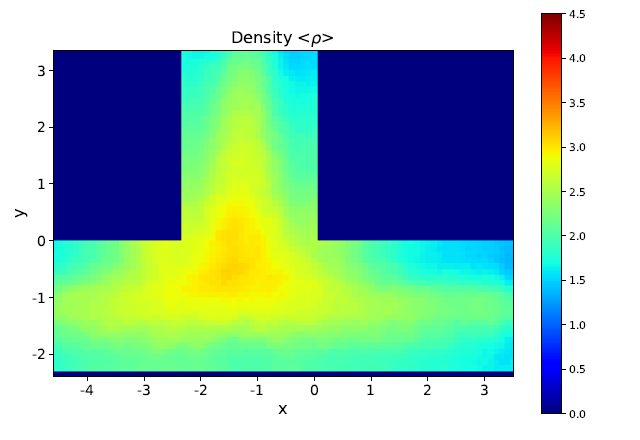}
			\caption{Simulated data set: static floor field with obstacle density $0.3$.}
		\end{subfigure}
		\begin{subfigure}{0.49\linewidth}
			\centering
			\includegraphics[width=\textwidth]{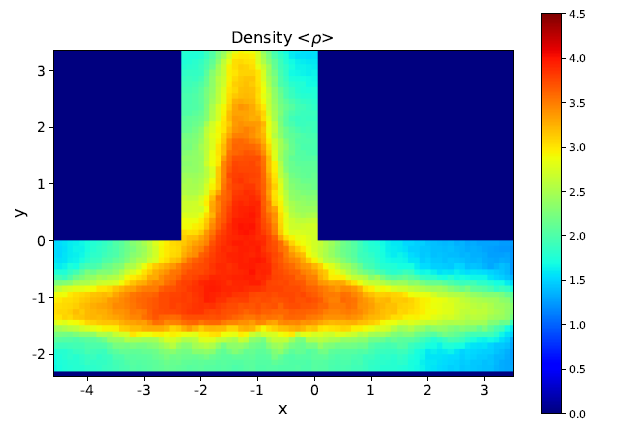}
			\caption{Simulated data set: static floor field with obstacle density $0.9$.}
		\end{subfigure}
		\caption{Average Voronoi density for the experimental data set (a) compared to the simulation with $4$ different navigation fields (b-e).}
	\end{figure}

We found the best agreement between simulated data and the experimental data for the static floor field with obstacle density with factor $0.3$. For the following simulations, this configuration is used. 

\subsubsection{Calculation density heatmaps}

Analogue to \cite{goedel-2018}, the density heatmaps are calculated from the pedestrian trajectories with a Gaussian density function \cite{seitz-2012}. 

\begin{figure}[H]
	\centering
\begin{subfigure}{0.49\linewidth}
	\centering
	\includegraphics[width=\textwidth]{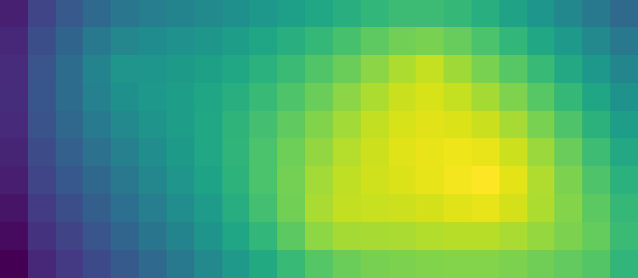}
	\caption{Experimental dataset: $40\%$ of participants come from the left origin, $60\%$ come from the right origin.}
\end{subfigure} 
\begin{subfigure}{0.49\linewidth}
	\centering
	\includegraphics[width=\textwidth]{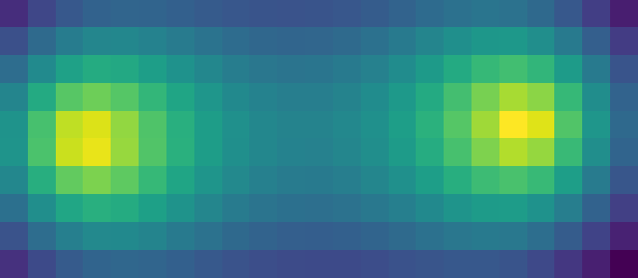}
	\caption{Experimental dataset: $50\%$ of participants come from the left origin, $50\%$ come from the right origin.}
	\end{subfigure}
\begin{subfigure}{0.49\linewidth}
	\centering
	\includegraphics[width=\textwidth]{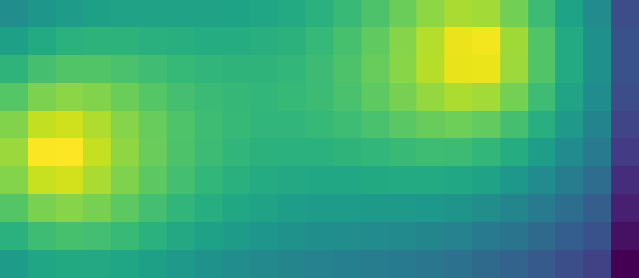}
	\caption{Simulated dataset: $0\%$ of agents come from the left origin, $100\%$ come from the right origin.}
	\end{subfigure}
\begin{subfigure}{0.49\linewidth}
	\centering
	\includegraphics[width=\textwidth]{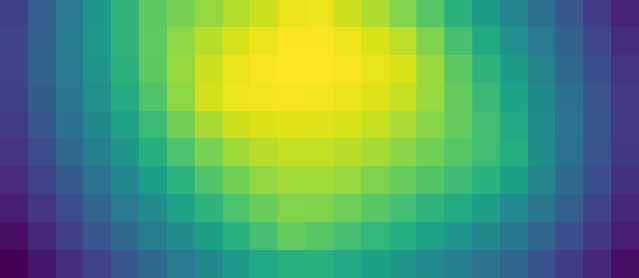}
	\caption{Simulated dataset: $17\%$ of agents come from the left origin, $83\%$ come from the right origin.}
	\end{subfigure}
	\caption{Exemplary density heatmaps for both experimental and simulated datasets.}
\end{figure}

\subsubsection{Comparison of simulation and experiment}

\begin{figure}[H]
	\centering
		\begin{subfigure}{0.49\linewidth}
		\centering
		\includegraphics[width=\textwidth]{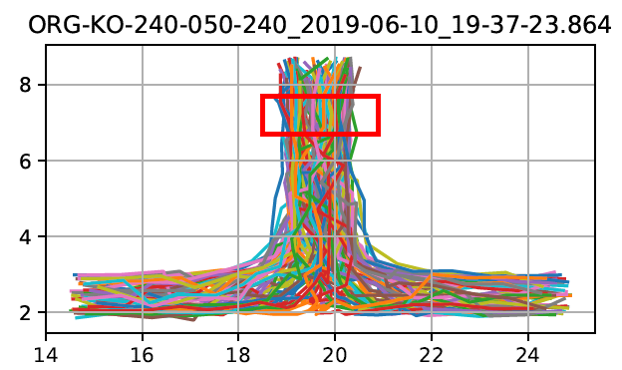}
		\caption{Trajectories of $134$ participants in the experiment ($b_{\text{entrance}}=0.5\,\text{m}$).}
	\end{subfigure}
	\begin{subfigure}{0.49\linewidth}
		\centering
		\includegraphics[width=\textwidth]{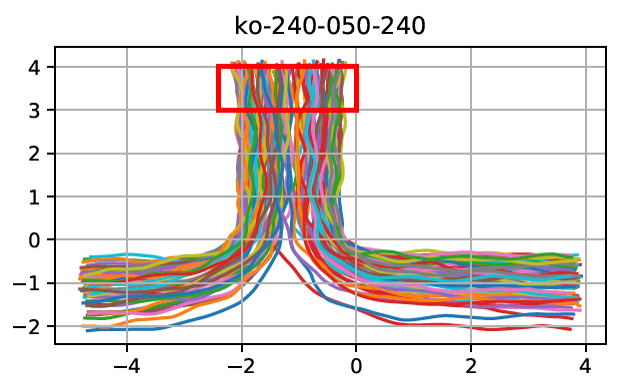}
		\caption{Trajectories of a simulation with  $134$ agents ($b_{\text{entrance}}=0.5\,\text{m}$).}
	\end{subfigure}
	\begin{subfigure}{0.49\linewidth}
	\centering
	\includegraphics[width=\textwidth]{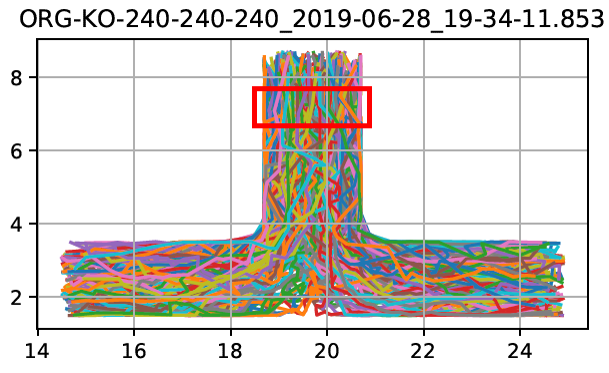}
	\caption{Trajectories of $303$ participants in the experiment ($b_{\text{entrance}}=2.40\,\text{m}$).}
	\end{subfigure}
	\begin{subfigure}{0.49\linewidth}
		\centering
		\includegraphics[width=\textwidth]{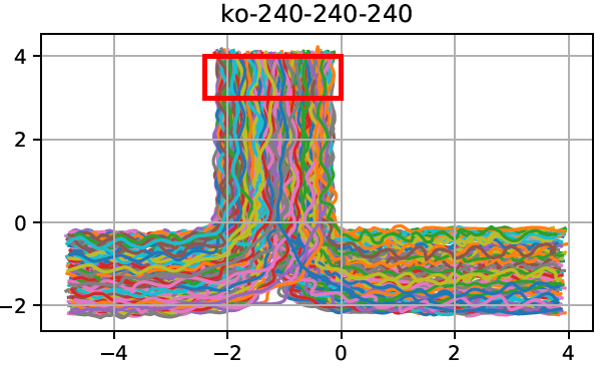}
		\caption{Trajectories of a simulation with  $303$ agents ($b_{\text{entrance}}=2.40\,\text{m}$).	}
	\end{subfigure}
	\caption{Comparison of trajectories between experiment and simulation.}
	\label{fig:traj}
\end{figure}

We compare the obtained simulated data with the experimental dataset. In Figure \ref{fig:traj}, the trajectories for two configurations (entrance widths $0.5\,m$ and $2.4\,m$) are compared. We observe that the agent flows / trajectories from left and right origin are more mixed in the simulation compared to the pedestrian trajectories from the experiment. For the experimental data, we observe an area free of pedestrians below the corridor. This effect is stronger for smaller entrance widths. We can mainly observe the effect in the experimental data set, only slightly in the simulated data. 

\begin{figure}[H]
	\centering
	\includegraphics[width=\textwidth]{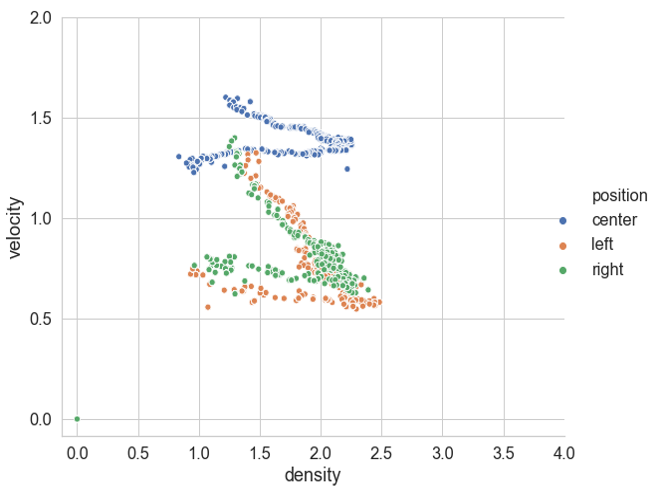}
\caption{Fundamental diagrams for simulated data in the T-junction setup. The simulation was set up according to the experiment in \cite{zhang-2011} with $303$ agents and an entrance width of $2.4\,m$.}
\label{fig:FD}
\end{figure}

In addition, we show the fundamental diagrams for the simulated data. It can be compared to the results from the experiment in \cite{zhang-2011}. In Figure \ref{fig:FD}, the density-speed relationship is evaluated for the three measurement areas defined in \cite{boltes-2011} that are depicted in red in Figure \ref{fig:simulation}.

\subsubsection{Simulated dataset}

Our simulated data set consists of three simulation runs for each of the seven scenarios (240-50-240, 240-60-240, 240-80-240, 240-100-240, 240-120-240, 240-150-240, 240-240-240), analogue to \cite{boltes-2011}. The tuple of numbers refers to the size of the entrance of the waiting area $b_{\text{entrance}}$, the width of the corridor $b_{\text{cor1}}$ and the width of the corridor towards the exit $b_{\text{exit}}$. In total, we obtain $14.714$ samples from the simulations. 

Due to the functional whiteness of the simulator, we may have many identical density heatmaps. Since this increases the noise in the relationship between input and output, we decided to throw out identical consecutive heatmaps.

After removing duplicates, we end up with $7.895$ heatmaps. When looking closer at Table \ref{tab:rf-vadere-dist}, we notice the heatmaps are not equally distributed between the existing origin distributions and the majority of heatmaps correspond to a $50\%\mid50\%$ distribution. To prevent the random forest model to only predicts $50\%\mid50\%$, we reduce the number of heatmaps for the equal case, see Table \ref{tab:rf-vadere-dist2}. In the end, $7.023$ heatmaps remain.

\begin{minipage}{0.49\textwidth}
	\begin{table}[H]
		\caption{Origin distributions in full dataset.}
		\label{tab:rf-vadere-dist}
		\begin{tabular}{ccc}
			\toprule
			{Left} &     {Right} &  {\# Heatmaps} \\
			\midrule
			0.000000 &  1.000000 &     557 \\
			0.125000 &  0.875000 &       2 \\
			0.142857 &  0.857143 &       7 \\
			0.166667 &  0.833333 &      70 \\
			0.200000 &  0.800000 &     245 \\
			0.222222 &  0.777778 &       1 \\
			0.250000 &  0.750000 &     444 \\
			0.285714 &  0.714286 &      64 \\
			0.333333 &  0.666667 &     853 \\
			0.375000 &  0.625000 &      23 \\
			0.400000 &  0.600000 &     592 \\
			0.428571 &  0.571429 &     154 \\
			0.444444 &  0.555556 &       3 \\
			0.500000 &  0.500000 &    1759 \\
			0.555556 &  0.444444 &       4 \\
			0.571429 &  0.428571 &     111 \\
			0.600000 &  0.400000 &     652 \\
			0.625000 &  0.375000 &      22 \\
			0.666667 &  0.333333 &     931 \\
			0.714286 &  0.285714 &      54 \\
			0.750000 &  0.250000 &     469 \\
			0.800000 &  0.200000 &     220 \\
			0.833333 &  0.166667 &      57 \\
			0.857143 &  0.142857 &      11 \\
			0.875000 &  0.125000 &       4 \\
			1.000000 &  0.000000 &     586 \\
			\bottomrule
		\end{tabular}
	\end{table}
\end{minipage}
\begin{minipage}{0.49\textwidth}
	\begin{table}[H]
		\caption{Origin distributions in reduced dataset.}
		\label{tab:rf-vadere-dist2}
		\begin{tabular}{ccc}
			\toprule
			{Left} &     {Right} &  {\# Heatmaps} \\
			\midrule
			0.000000 &  1.000000 &     557 \\
			0.125000 &  0.875000 &       2 \\
			0.142857 &  0.857143 &       7 \\
			0.166667 &  0.833333 &      70 \\
			0.200000 &  0.800000 &     245 \\
			0.222222 &  0.777778 &       1 \\
			0.250000 &  0.750000 &     444 \\
			0.285714 &  0.714286 &      64 \\
			0.333333 &  0.666667 &     853 \\
			0.375000 &  0.625000 &      23 \\
			0.400000 &  0.600000 &     592 \\
			0.428571 &  0.571429 &     154 \\
			0.444444 &  0.555556 &       3 \\
			0.500000 &  0.500000 &    \bf{ \color{red}{887} }\\
			0.555556 &  0.444444 &       4 \\
			0.571429 &  0.428571 &     111 \\
			0.600000 &  0.400000 &     652 \\
			0.625000 &  0.375000 &      22 \\
			0.666667 &  0.333333 &     931 \\
			0.714286 &  0.285714 &      54 \\
			0.750000 &  0.250000 &     469 \\
			0.800000 &  0.200000 &     220 \\
			0.833333 &  0.166667 &      57 \\
			0.857143 &  0.142857 &      11 \\
			0.875000 &  0.125000 &       4 \\
			1.000000 &  0.000000 &     586 \\
			\bottomrule
		\end{tabular}
	\end{table}
\end{minipage}

\subsection{Random forest}

We decided to use random forest as machine learning model as in \cite{goedel-2018}. It is an easy and robust algorithm that needs only a few parameters and often provides good results. Consequently, the configuration is easy, and the model can be used right away. In addition, the model is robust on the parameters which is not necessarily the case for machine learning models. Furthermore, the output is interpretable. 

The input to our random forest routine is a single density heatmap, generated as described before, from pedestrian positions using a Gaussian function. We want to predict the percentage of pedestrians stemming from each of the two origins from the density heatmap. 

For the random forest we use the \texttt{RandomForestRegressor} from the Python library scikit-learn \cite{pedregosa-2011}. Before training, we shuffle the heatmaps and divide them into training and test sets with $80\%$ and $20\%$ respectively. For each origin, we train one forest with $50$ trees. Consequently, when evaluating the model, each origin is also predicted separately,  and the predictions are normalized over both directions to assure that they sum up to $100\%$. 


\section{Results and discussion}
Now, we use our Random forest setup to predict the origin distributions from density heatmaps. We use three different data set: First, the dataset that contains simulations performed with Vadere. Second, the experimental dataset calculated from the trajectory data. Third, a hybrid dataset where we train the random forest with simulated data but evaluate its performance on the experimental data. For each setup, we perform five runs to study the variation in the results.

\subsection{Performance measure}
The random forest routines can provide an out of bag error estimate as well as the coefficient of determination. However, both quantities are applied before the model predictions are normalized. Therefore, we split our dataset into a training and a test set and compare the prediction on the test set to the corresponding response to evaluate the model. In a first step, we evaluate the Euclidean norm of the difference vector. In the second step, we derive a relative error. Since we normalize the prediction, the maximum error
\begin{equation}
e_{max} = \sqrt{2 \cdot 100^2} \approx 141.42
\end{equation}
occurs if we predict that all pedestrians within the cutout head left while, in fact, they head right (or straight). The relative error is then
\begin{equation}
e = 100 \cdot \frac{y - \hat{y}}{e_{max}} , 
\end{equation}
where $y$ is the response on the test set and $\hat{y}$ is the prediction on the test set. Thereby we obtain a relative error of the prediction that is easy to interpret. Thus, the relative error is used as a measure.

\subsection{Retracing using simulated data}
\label{sec:resultsSimulation}

For the first setup, we train and test the random forest solely on simulated data. We generate simulation data for all distributions of pedestrians on the origins. We expect an accuracy of  roughly $80$\%, similar to our observations in \cite{goedel-2018} where we predicted the target distributions. The results in Table \ref{tab:rf-vadere-result} show an accuracy of about $80\%$ of the setup with a standard deviation of about $15\%$.

\begin{table}[H]
	\centering
	\caption{Relative error of  random forest prediction using simulated data ($7.023$ density heatmaps).}
	\label{tab:rf-vadere-result}
	\begin{tabular}{lccccc}
		\toprule
		{}  & {Run 1} &  {Run 2} &  {Run 3} &  {Run 4} &  {Run 5} \\
		\midrule
		Mean Euclidean error & 19.23\% & 20.15\% & 19.85\% & 19.78\% & 19.57\% \\
		Stdev  Euclidean error & 14.98\% & 15.08\% & 15.93\% & 15.47\% & 15.29\% \\
		\bottomrule
	\end{tabular}
\end{table} 

\subsection{Retracing using experimental data}

In this section, we analyze the results of the random forest prediction based on experimental data. The dataset from the experiments shows only a few  different distributions of the pedestrians on the origins. 
In experimental data set, mainly an equal distribution of the pedestrians on the origins is observed. We mainly observe the following distributions: $100\mid0$, $50\mid50$, $66\mid34$, $75\mid25$. This is a consequence of the experimental setup and the size of the observation area. 
Table \ref{tab:rf-experiment-result} summarizes the results for the experimental dataset for both training and evaluation. The prediction accuracy is lower than when we used solely simulated data. We observe that the equal distribution is estimated well, due to the large data base. Other distributions are estimated with lower accuracy. Most of the predicted distributions are targets from the training set even though we use a regressor, not a classifier. 

\begin{table}[H]
	\centering
	\caption{Relative error of  random forest prediction using the experiment data ($6.678$ density heatmaps).}
	\label{tab:rf-experiment-result}
	\begin{tabular}{lccccc}
		\toprule
		{}  & {Run 1} &  {Run 2} &  {Run 3} &  {Run 4} &  {Run 5} \\
		\midrule
		Mean Euclidean error & 11.12\% & 9.68\% & 9.66\% & 10.33\% & 9.82\% \\
		Stdev  Euclidean error & 15.89\% & 13.46\% & 13.19\% & 14.48\% & 13.89\% \\
		\bottomrule
	\end{tabular}
\end{table} 

\subsection{Retracing using hybrid approach}

We observed that the available experimental data has only a few different distributions. Unfortunately, performing experiments is expensive, time-consuming, and complicated. Ideally, we would have a heterogeneous set of participants to reflect the average population well. In addition, results may strongly depend on the setup, how the subjects are primed, if and how subjects know each other, their form of the day and many more. In addition, the culture may have an impact on the behavior, so we need comparable studies. 

Simulated data on the other hand can be generated easily and cheap. For applications, none or only limited experimental data is available. 

As a result, we decided to complement the experimental data with simulated data: The algorithm is trained solely on simulated data. The testing is performed in two steps: 
First, testing of the trained random forest on the simulated data. This aims to find out how well the forest is trained. The results should be identical to Section \ref{sec:resultsSimulation}. Second, testing of the trained random forest on the experimental data. The goal of this part is to find out if the gained knowledge from the simulations can be extrapolated to real data. 

Table \ref{tab:rf-hybrid-result} lists the results for $5$ trainings of the random forest algorithm. As the input data is shuffled, the separation in training and test data varies among the runs. We observe a mean error about $30\%$ with a standard deviation of approximately $25\%$. As expected, the results are inferior to training and testing with the same data set. Both the size of the error and the size of the standard deviation have almost doubled. Nevertheless, we still obtain an accuracy of about $70\%$. That means, it is possible to train with simulated data and use the trained forest for predictions on experimental data. 

\begin{table}[H]
	\centering
	\caption{Relative error of  random forest prediction using the hybrid dataset. }
	\label{tab:rf-hybrid-result}
	\begin{tabular}{lccccc}
		\toprule
		{}  & {Run 1} &  {Run 2} &  {Run 3} &  {Run 4} &  {Run 5} \\
		\midrule
		Mean Euclidean error & 29.36\% & 29.39\% & 29.24\% & 29.43\% & 29.40\% \\
		Stdev  Euclidean error & 24.48\% & 24.73\% & 24.51\% & 24.70\% & 24.80\% \\
		\bottomrule
	\end{tabular}
\end{table} 

\section{Extending the observation area}

In the previous chapters, we saw limitations of the results due to the low variation of the origin distributions present in the data sets. Therefore, we decided to use a larger observation area so that more than a maximum of two pedestrians can be present at one time. That allows a larger variation in the distributions on the origins. The new observation area has the size of $2.4\,\text{m} \times 2\,\text{m} (B \times H)$. It is again placed over the full corridor width just before the target. 

\begin{figure}[H]
	\centering
	\begin{subfigure}{0.49\linewidth}
		\centering	\includegraphics[width=\textwidth]{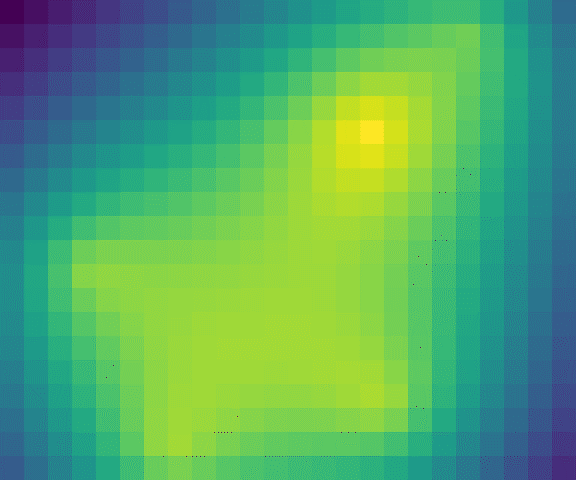}
		\caption{Experimental dataset: $60\%$ of participants come from the left origin, $40\%$ come from the right origin.}
	\end{subfigure}
	\begin{subfigure}{0.49\linewidth}
		\centering
		\includegraphics[width=\textwidth]{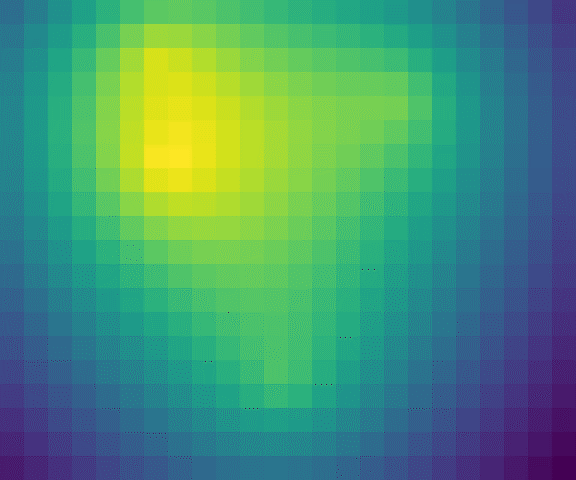}
		\caption{Simulated dataset: $25\%$ of agents come from the left origin, $75\%$ come from the right origin.}
	\end{subfigure}
	\caption{Exemplary density heatmaps for the larger observation area (\#2).}
	\label{fig:heatmapObsArea2}
\end{figure}

In Figure \ref{fig:heatmapObsArea2}, we depicted two exemplary heatmaps for the larger observation area. Due to the larger observation area (2), more variation in the data is seen. Nevertheless, the equal distribution on both origins ($50\%\mid 50\%$) and distributions close to it are still most common.

In Tables \ref{tab:rf-vadere-result-2}, \ref{tab:rf-experiment-result-2}, \ref{tab:rf-hybrid-result-2} the results for the random forest prediction of the origin distributions are shown for solely simulated data, solely experimental data, and the hybrid dataset with simulated data for training and experimental data for testing, respectively. The results for all datasets have improved compared to the smaller observation area (\#1). The improvement can be seen from a decrease in the mean error in combination with a decrease of the standard deviation. That means, that there is less variation of the quality of the prediction within the test set. 

\begin{table}[H]
	\centering
	\caption{Relative error of  random forest prediction using simulated data generated with Vadere (observation area \#2, $9.666$ samples).}
	\label{tab:rf-vadere-result-2}
	\begin{tabular}{lccccc}
		\toprule
		{}  & {Run 1} &  {Run 2} &  {Run 3} &  {Run 4} &  {Run 5} \\
		\midrule
		Mean Euclidean error & 12.02\% & 12.03\% & 12.09\% & 11.61\% & 11.82\% \\
		Stdev  Euclidean error & 9.64\% & 10.07\% & 9.82\% & 9.56\% & 9.39\% \\
		\bottomrule
	\end{tabular}
\end{table}

\begin{table}[H]
	\centering
	\caption{Relative error of random forest prediction using experiment data (observation area \#2, $7.221$ samples).}
	\label{tab:rf-experiment-result-2}
	\begin{tabular}{lccccc}
		\toprule
		{}  & {Run 1} &  {Run 2} &  {Run 3} &  {Run 4} &  {Run 5} \\
		\midrule
		Mean Euclidean error & 7.65\% & 7.63\% & 7.70\% & 7.85\% & 8.02\% \\
		Stdev  Euclidean error & 9.33\% & 9.27\% & 8.37\% & 8.88\% & 9.55\% \\
		\bottomrule
	\end{tabular}
\end{table} 

\begin{table}[H]
	\centering
	\caption{Relative error of  random forest prediction using the hybrid dataset (observation area \#2).}
	\label{tab:rf-hybrid-result-2}
	\begin{tabular}{lccccc}
		\toprule
		{}  & {Run 1} &  {Run 2} &  {Run 3} &  {Run 4} &  {Run 5} \\
		\midrule
		Mean Euclidean error & 19.78\% & 19.65\% & 19.67\% & 19.80\% & 19.56\% \\
		Stdev  Euclidean error & 18.63\% & 18.42\% & 18.31\% & 18.77\% & 18.15\% \\
		\bottomrule
	\end{tabular}
\end{table} 

\section{Conclusion and outlook}
We were able to retrace the origin distributions based on single density heatmap. Basis of this study was a merging experiment (T-junction) where two pedestrian streams head for the same destination. We trained as random forest regressor with three different data sets: The first data set stems from simulations with the simulation framework Vadere. The setup of the simulation resembles the T-junction experiment. The second dataset is the publicly available data of the experiment. The last dataset is a hybrid dataset of both sources that means we used simulated data for training the random forest and experimental data for testing. 

We observed the best results ($\approx 92\%$) on experimental data which is most likely  due to the limited set of origin distributions in the data set. The accuracy of the random forest model on simulated data was at about $88\%$. Finally, the hybrid approach yielded an accuracy of approximately $80\%$. This is a promising result that motivates the application of the approach on real data. Especially for applications, we believe the results of the hybrid approach are encouraging because often  there will  not be enough training data available. 

In the future, we would like to look into other scenarios and topographies more complicated than the one studied here. Unfortunately, there is only limited amount of public data available. In addition, we want to try other learning models to see how they perform on the problem, in particular convolutional neural networks that are designed to work on image data. Furthermore, we believe that for more complicated scenarios, it will be beneficial or maybe even necessary to consider a time-dependent input, that is a time series of consecutive heatmaps.

\funding{This research was funded by the German Federal Ministry of Education and Research (BMBF) grant number 13N14464 (S2UCRE).}	

\appendix
\section[Observed origin distributions]{Origin distributions observed in different datasets (simulated data, experimental data) for both observation areas (1,2)}

\begin{table}[H]
	\caption{Origin distributions in experiment dataset using observation area \#1 ($2.4\,m \times 1\,m$).}
	\centering
	\begin{tabular}{lrrr}
		\toprule
		{} &     Left &  Right &  \#Frames \\
		\midrule
		0  &  0.000000 &  1.000000 &    1372 \\
		1  &  0.166667 &  0.833333 &      11 \\
		2  &  0.200000 &  0.800000 &      45 \\
		3  &  0.250000 &  0.750000 &     294 \\
		4  &  0.285714 &  0.714286 &       3 \\
		5  &  0.333333 &  0.666667 &     881 \\
		6  &  0.400000 &  0.600000 &     517 \\
		7  &  0.428571 &  0.571429 &      20 \\
		8  &  0.500000 &  0.500000 &    2241 \\
		9  &  0.571429 &  0.428571 &      23 \\
		10 &  0.600000 &  0.400000 &     471 \\
		11 &  0.625000 &  0.375000 &       2 \\
		12 &  0.666667 &  0.333333 &     838 \\
		13 &  0.714286 &  0.285714 &       2 \\
		14 &  0.750000 &  0.250000 &     304 \\
		15 &  0.800000 &  0.200000 &      35 \\
		16 &  1.000000 &  0.000000 &     738 \\
		\bottomrule
	\end{tabular}
\end{table}

\begin{table}[H]
	\caption{Origin distributions in experiment dataset using observation area \#2 ($2.4\,\text{m} \times 2\,\text{m}$).}
	\centering
\begin{tabular}{lrrr}
	\toprule
		{} &     Left &  Right &  \#Frames \\
	\midrule
	0  &  0.000000 &  1.000000 &     873 \\
	1  &  0.125000 &  0.875000 &       1 \\
	2  &  0.142857 &  0.857143 &       6 \\
	3  &  0.166667 &  0.833333 &      16 \\
	4  &  0.200000 &  0.800000 &      45 \\
	5  &  0.222222 &  0.777778 &      15 \\
	6  &  0.250000 &  0.750000 &     166 \\
	7  &  0.285714 &  0.714286 &     164 \\
	8  &  0.300000 &  0.700000 &      25 \\
	9  &  0.333333 &  0.666667 &     471 \\
	10 &  0.363636 &  0.636364 &      62 \\
	11 &  0.375000 &  0.625000 &     347 \\
	12 &  0.400000 &  0.600000 &     458 \\
	13 &  0.416667 &  0.583333 &      22 \\
	14 &  0.428571 &  0.571429 &     473 \\
	15 &  0.444444 &  0.555556 &     478 \\
	16 &  0.454545 &  0.545455 &      81 \\
	17 &  0.461538 &  0.538462 &       1 \\
	18 &  0.500000 &  0.500000 &     829 \\
	19 &  0.538462 &  0.461538 &       1 \\
	20 &  0.545455 &  0.454545 &      76 \\
	21 &  0.555556 &  0.444444 &     381 \\
	22 &  0.571429 &  0.428571 &     386 \\
	23 &  0.583333 &  0.416667 &      22 \\
	24 &  0.600000 &  0.400000 &     461 \\
	25 &  0.615385 &  0.384615 &       8 \\
	26 &  0.625000 &  0.375000 &     200 \\
	27 &  0.636364 &  0.363636 &      38 \\
	28 &  0.666667 &  0.333333 &     504 \\
	29 &  0.700000 &  0.300000 &      20 \\
	30 &  0.714286 &  0.285714 &      55 \\
	31 &  0.750000 &  0.250000 &     221 \\
	32 &  0.777778 &  0.222222 &       1 \\
	33 &  0.800000 &  0.200000 &      72 \\
	34 &  0.833333 &  0.166667 &      27 \\
	35 &  1.000000 &  0.000000 &     215 \\
	\bottomrule
\end{tabular}
\end{table}

\begin{table}[H]
	\caption{Origin distributions observed in simulated dataset using observation area \#1 ($2.4\,\text{m} \times 1\,\text{m}$).}
	\centering
\begin{tabular}{lrrr}

	\toprule
		{} &     Left &  Right &  \#Frames \\
	\midrule
	0  &  0.000000 &  1.000000 &     557 \\
	1  &  0.125000 &  0.875000 &       2 \\
	2  &  0.142857 &  0.857143 &       7 \\
	3  &  0.166667 &  0.833333 &      70 \\
	4  &  0.200000 &  0.800000 &     245 \\
	5  &  0.222222 &  0.777778 &       1 \\
	6  &  0.250000 &  0.750000 &     444 \\
	7  &  0.285714 &  0.714286 &      64 \\
	8  &  0.333333 &  0.666667 &     853 \\
	9  &  0.375000 &  0.625000 &      23 \\
	10 &  0.400000 &  0.600000 &     592 \\
	11 &  0.428571 &  0.571429 &     154 \\
	12 &  0.444444 &  0.555556 &       3 \\
	13 &  0.500000 &  0.500000 &     887 \\
	14 &  0.555556 &  0.444444 &       4 \\
	15 &  0.571429 &  0.428571 &     111 \\
	16 &  0.600000 &  0.400000 &     652 \\
	17 &  0.625000 &  0.375000 &      22 \\
	18 &  0.666667 &  0.333333 &     931 \\
	19 &  0.714286 &  0.285714 &      54 \\
	20 &  0.750000 &  0.250000 &     469 \\
	21 &  0.800000 &  0.200000 &     220 \\
	22 &  0.833333 &  0.166667 &      57 \\
	23 &  0.857143 &  0.142857 &      11 \\
	24 &  0.875000 &  0.125000 &       4 \\
	25 &  1.000000 &  0.000000 &     586 \\
	\bottomrule
\end{tabular}
\end{table}
\newpage
\begin{table}[H]
	\caption{Origin distributions in simulated dataset using observation area \#2 ($2.4\,\text{m} \times 2\,\text{m}$).}
	\centering
	\begin{minipage}{0.49\textwidth}
	\begin{tabular}{lrrr }
		\toprule
		{} &     Left &  Right &  \#Frames \\
		\midrule
		0  &  0.000000 &  1.000000 &      70 \\
		1  &  0.100000 &  0.900000 &       5 \\
		2  &  0.111111 &  0.888889 &       6 \\
		3  &  0.125000 &  0.875000 &      14 \\
		4  &  0.142857 &  0.857143 &      14 \\
		5  &  0.166667 &  0.833333 &      67 \\
		6  &  0.181818 &  0.818182 &      18 \\
		7  &  0.200000 &  0.800000 &      89 \\
		8  &  0.222222 &  0.777778 &      45 \\
		9  &  0.230769 &  0.769231 &       6 \\
		10 &  0.250000 &  0.750000 &     179 \\
		11 &  0.272727 &  0.727273 &      79 \\
		12 &  0.285714 &  0.714286 &     122 \\
		13 &  0.300000 &  0.700000 &     126 \\
		14 &  0.307692 &  0.692308 &      31 \\
		15 &  0.333333 &  0.666667 &     565 \\
		16 &  0.357143 &  0.642857 &      37 \\
		17 &  0.363636 &  0.636364 &     237 \\
		18 &  0.375000 &  0.625000 &     246 \\
		19 &  0.384615 &  0.615385 &     138 \\
		20 &  0.400000 &  0.600000 &     612 \\
		21 &  0.416667 &  0.583333 &     295 \\
		22 &  0.428571 &  0.571429 &     431 \\
		23 &  0.437500 &  0.562500 &       3 \\
		24 &  0.444444 &  0.555556 &     326 \\
		25 &  0.454545 &  0.545455 &     405 \\
		26 &  0.461538 &  0.538462 &     228 \\
		27 &  0.466667 &  0.533333 &      23 \\
		28 &  0.470588 &  0.529412 &       2 \\
		\bottomrule
	\end{tabular}
\end{minipage}
\begin{minipage}{0.48\textwidth}
	\begin{tabular}{lrrr }
	\toprule
		{} &     Left &  Right &  \#Frames \\
	\midrule
	28 &  0.470588 &  0.529412 &       2 \\
	29 &  0.500000 &  0.500000 &     920 \\
	30 &  0.533333 &  0.466667 &      17 \\
	31 &  0.538462 &  0.461538 &     187 \\
	32 &  0.545455 &  0.454545 &     383 \\
	33 &  0.555556 &  0.444444 &     285 \\
	34 &  0.562500 &  0.437500 &       3 \\
	35 &  0.571429 &  0.428571 &     482 \\
	36 &  0.583333 &  0.416667 &     270 \\
	37 &  0.600000 &  0.400000 &     548 \\
	38 &  0.615385 &  0.384615 &     133 \\
	39 &  0.625000 &  0.375000 &     264 \\
	40 &  0.636364 &  0.363636 &     204 \\
	41 &  0.642857 &  0.357143 &      36 \\
	42 &  0.666667 &  0.333333 &     594 \\
	43 &  0.692308 &  0.307692 &      31 \\
	44 &  0.700000 &  0.300000 &      98 \\
	45 &  0.714286 &  0.285714 &     178 \\
	46 &  0.727273 &  0.272727 &      64 \\
	47 &  0.733333 &  0.266667 &       1 \\
	48 &  0.750000 &  0.250000 &     185 \\
	49 &  0.769231 &  0.230769 &       4 \\
	50 &  0.777778 &  0.222222 &      35 \\
	51 &  0.800000 &  0.200000 &     102 \\
	52 &  0.833333 &  0.166667 &      93 \\
	53 &  0.857143 &  0.142857 &      39 \\
	54 &  0.875000 &  0.125000 &       8 \\
	55 &  0.888889 &  0.111111 &       2 \\
	56 &  1.000000 &  0.000000 &      81 \\
	\bottomrule
\end{tabular}
\end{minipage}
\end{table}

\newpage
\bibliographystyle{plain}
\bibliography{Lit_short}

\end{document}